\documentclass[conference]{IEEEtran}
\IEEEoverridecommandlockouts
\usepackage{cite}
\usepackage{stfloats}
\usepackage{amsmath,amssymb,amsfonts}
\usepackage{algorithmic}
\usepackage{graphicx}
\usepackage{textcomp}
\usepackage{caption}
\usepackage{tabularx}
\usepackage{longtable}
\usepackage{array}
\usepackage[table]{xcolor}
\usepackage{float}
\usepackage{ragged2e}
\usepackage{subcaption}
\usepackage{comment}

\def\BibTeX{{\rm B\kern-.05em{\sc i\kern-.025em b}\kern-.08em
    T\kern-.1667em\lower.7ex\hbox{E}\kern-.125emX}}
\begin{document}

\title{HySAFE-AI: Hybrid Safety Architectural Analysis Framework for AI Systems: A Case Study\\}

\author{\IEEEauthorblockN{1\textsuperscript{st} Mandar Pitale}
\IEEEauthorblockA{\textit{Nvidia Corporation}\\
mpitale@nvidia.com}
\and
\IEEEauthorblockN{2\textsuperscript{nd} Jelena Frtunikj}
\IEEEauthorblockA{\textit{Nvidia GmbH}\\
jfrtunikj@nvidia.com}
\and
\IEEEauthorblockN{3\textsuperscript{rd} Abhinaw Priyadershi}
\IEEEauthorblockA{\textit{Nvidia Corporation}\\
apriyadershi@nvidia.com}
\and
\IEEEauthorblockN{4\textsuperscript{th} Vasu Singh}
\IEEEauthorblockA{\textit{Nvidia Corporation}\\
vasus@nvidia.com}
\and
\IEEEauthorblockN{5\textsuperscript{th} Maria Spence}
\IEEEauthorblockA{\textit{Nvidia Corporation}\\
mspence@nvidia.com}

}

\maketitle

\begin{abstract}
AI has become integral to safety-critical areas like autonomous driving systems (ADS) and robotics. The architecture of recent autonomous systems are trending toward end-to-end (E2E) monolithic architectures such as large language models (LLMs) and vision language models (VLMs). In this paper, we review different architectural solutions and then evaluate the efficacy of common safety analyses such as failure modes and effect analysis (FMEA) and fault tree analysis (FTA). We show how these techniques can be improved for the intricate nature of the foundational models, particularly in how they form and utilize latent representations. We introduce HySAFE-AI, Hybrid Safety Architectural Analysis Framework for AI Systems, a hybrid framework that adapts traditional methods to evaluate the safety of AI systems. Lastly, we offer hints of future work and suggestions to guide the evolution of future AI safety standards.
\end{abstract}


\section{Introduction}
 
Artificial intelligence (AI) integration into safety-critical domains like autonomous driving, healthcare, and industrial robotics offers significant improvements in efficiency and decision-making. However, the complexity and opacity of foundation models (FMs), including LLMs and VLMs~\cite{Wayve2024}~\cite{Wayve2025}~\cite{Yang2023A}, have introduced new safety challenges due to their closed-box architectures and dynamic behaviors that are inherently difficult to analyze. 

The widespread deployment of FMs has highlighted significant gaps in ensuring safe and reliable operation. Unlike traditional domain-specific AI models, FMs operate with limited interpretability, making it challenging to trace decision-making processes and identify potential failure points. This lack of transparency raises concerns about their suitability for safety-critical scenarios.

\noindent \textbf{Evolution of AI System Architecture for Autonomous Cars.}
Autonomous driving systems traditionally use \emph{modular architectures} with distinct components for perception, prediction, planning, and control. Although offering interpretability benefits, modular architectures encounter inherent limitations that impede robustness for the following reasons: (1) the complexity of managing the interfaces and interactions between these independent modules, (2) cascading errors across modules~\cite{pitale2024}.

\emph{End-to-end (E2E) models}~\cite{Wayve2024,Wayve2025,Yang2023A} address these limitations by directly mapping sensor data to control commands through unified, fully differentiable architectures trained on extensive driving data. E2E systems simplify development and reduce component integration challenges but introduce significant interpretability concerns due to their ``closed-box'' nature, making failure analysis and safety assurance difficult.

\emph{Hybrid approaches} combine modular and E2E benefits by retaining identifiable modules within an E2E trainable framework~\cite{rosero2024hybrid}, enabling interpretability while minimizing information loss across components.

\noindent \textbf{Contributions.}
Safety assurance in safety-critical systems has traditionally relied on established analysis methods, including FMEA, FTA, Event Tree Analysis (ETA), System-Theoretic Process Analysis (STPA), and Critical Path Analysis (CPA). These methods have been effective in evaluating traditional engineering systems and domain-specific AI applications with well-understood architectures. However, FM-based AI systems pose major challenges due to their scale, complexity, opacity, and lack of structural transparency. This paper addresses FM-related safety challenges by (1) identifying the limitations of traditional safety analyses for E2E ADS, (2) augmenting these analyses with appropriate measures, incorporating principles and techniques tailored to FM-based systems, and (3) constructing a reference E2E architecture with sufficient details to enable meaningful analysis, followed by application of the adapted analyses.

\section{Reference E2E Architecture and Related work}

This section provides the details about the E2E reference architecture, conducts comparative analysis of AI model integration across the perception-to-control stack, and  presents the critical limitations in safety standards and safety analysis methods when applied to latent-space-driven architectures.

\noindent \textbf{Reference E2E Architecture.}
Generative autonomous driving models (GenAD~\cite{Yang2023B}, GAIA-1~\cite{Wayve2024}, GAIA-2~\cite{Wayve2025}) share a unified architectural blueprint with three core components:

\emph{Latent Space Modeling:} All models operate in compressed latent spaces for computational efficiency. GenAD uses frozen VAE encoders with UNet denoising, GAIA-1 discretizes multimodal inputs into token sequences, and GAIA-2 uses space-time factorized transformers with 32× spatial compression.

\emph{Generative Backbone:} Latent denoisers iteratively refine corrupted features. GenAD utilizes UNet with temporal reasoning blocks, while GAIA-1/2 use transformer-based world models - GAIA-1 with autoregressive discrete tokens and GAIA-2 with an 8.4B parameter continuous latent transformer.

\emph{Multimodal Conditioning:} Critical for scenario control and interpretability. GenAD performs text conditioning via BLIP-2/CLIP, GAIA uses semantic prompts for scene elements and ego trajectory conditioning.

\emph{Video Synthesis \& Inference:} GenAD generates video from denoised latents with MLP trajectory decoding, GAIA-1 uses diffusion decoders, and GAIA-2 reconstructs multi-camera videos via sliding windows. All support controllable scenario generation with varying capabilities.

\begin{figure}
\centering
\includegraphics[width=3.5in]{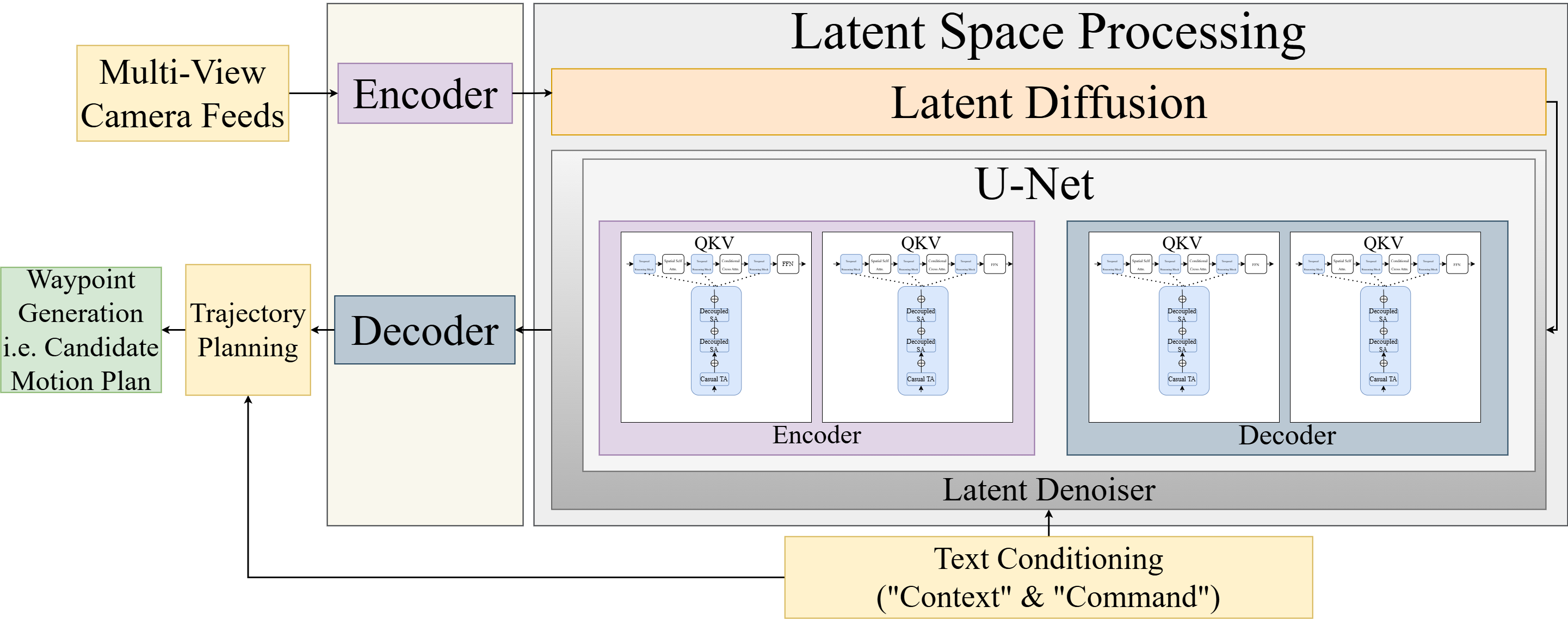}
\caption{Reference Architecture of Unified Framework} \label{fig2}
\end{figure}

The reference architecture in Figure~\ref{fig2} integrates GAIA-2~\cite{Wayve2025} and GenAD~\cite{Yang2023B} components for unified autonomous driving simulation and motion planning. Table~\ref{tab:components_features} details component functionality, interfaces, and features.

\begin{table}
    \centering
    {\scriptsize 
    \begin{tabular}{|>{\raggedright\arraybackslash}p{1.1cm}|>{\raggedright\arraybackslash}p{2.0cm}|>{\raggedright\arraybackslash}p{2.0cm}|>{\raggedright\arraybackslash}p{1.8cm}|}
        \hline
        \textbf{Component} & \textbf{Functionality} & \textbf{Interfaces} & \textbf{Key Features} \\ \hline
        Encoder & Compresses multi-view camera feeds into compact latent representations. Preserves spatiotemporal and semantic features. & Input: Raw camera frames (e.g., 448×960). \newline Output: Latent representations for diffusion/U-Net. & - High compression (32x spatial, 8x temporal \cite{Wayve2025}) \newline - Preserves spatiotemporal semantics. \\ \hline
      
        Latent Diffusion & Iteratively denoises latents to generate future states/scenes. Supports flow matching or diffusion (e.g., DDIM). & Input: Noisy latents or encoder outputs. Conditioning: Ego actions, text, 3D agents. \newline Output: Denoised latents & - Causal attention \cite{Yang2023B} \newline - Flow matching \cite{Wayve2025} \\ \hline
        
        U-Net & Spatial-temporal feature extraction as the backbone. Augmented with causal attention for video prediction & Input: Noisy latents and conditions \newline Output: Refined latents for decoding. & - Interleaved spatial/temporal blocks \cite{Yang2023B} \newline - Space-time factorization \cite{Wayve2025} \\ \hline
       
        Latent Denoiser & Refines noisy latents into clean representations (e.g., via DDIM). & Input: Noisy latents \newline Output: Final latent states & - Ensures temporal coherence \newline - Classifier-free guidance support \\ \hline
        
        Decoder & Reconstructs high-resolution video from latents. & Input: Denoised latents \newline Output: Photorealistic frames & - Multi-camera consistency \cite{Wayve2025} \newline - Frozen autoencoder \cite{Yang2023B} \\ \hline
     
        Trajectory Planner & Predicts future ego trajectories. & Input: Encoder/U-Net features \newline Output: Waypoints (e.g., 6 steps) & - Lightweight MLP \cite{Yang2023B} \newline - Kinematic action conditioning \cite{Wayve2025} \\ \hline
      
        Text Conditioning & Integrates language commands/metadata. & Input: Text prompts or labels \newline Output: Embedded conditions & - CLIP embeddings \cite{Yang2023B} \newline - Cross-attention \cite{Wayve2025} \newline - Negative Prompting \cite{Wayve2024} \\ \hline
    \end{tabular}
    }
    \captionsetup{skip=10pt}
    \caption{Architecture Components and Features}
    \label{tab:components_features}
\end{table}

NOTE: Components like U-Net or Latent Denoiser can be quantized to reduce memory and computational needs for deploying large models on resource-constrained automotive hardware. However, quantization may cause precision loss, requiring careful safety analysis to avoid performance degradation in safety-critical scenarios.

\noindent \textbf{Current Usage of AI Models in Autonomous Driving.} The autonomous driving stack incorporates diverse AI model architectures across Perception, Localization, Prediction, Planning, and Control layers. Traditional CNNs remain critical for real-time, safety-critical environments~\cite{bojarski2016end}. Foundation models (FMs)—large-scale, multi-modal AI systems—enable significant integration across perception, localization, prediction, and planning, enhancing generalization~\cite{AnImageisWorth}. Vision-Language Models (VLMs) like CLIP improve semantic understanding~\cite{radford2021learning}, while Large Language Models (LLMs) show emerging potential in reasoning and planning~\cite{shah2022lmnav}. The control layer remains predominantly reliant on legacy AI models to meet stringent latency and stability requirements.

\begin{table}
\centering
{\scriptsize 
\begin{tabular}{|>{\raggedright\arraybackslash}p{1.1cm}|>{\centering\arraybackslash}p{1.3cm}|>{\centering\arraybackslash}p{1.4cm}|>{\centering\arraybackslash}p{1.4cm}|>{\centering\arraybackslash}p{1.3cm}|}
\hline
\textbf{ADS Layer} & \textbf{Legacy AI} & \textbf{Foundation Models} & \textbf{Vision Language Models} & \textbf{Large Language Models} \\ \hline
Perception & \cellcolor{white} \cite{bojarski2016end}\cite{E2ELearningApproach}\cite{VisualBackProp} & \cellcolor{white} \cite{AnImageisWorth}\cite{E2EDetTransf}\cite{SurveyonOccupancy} & \cellcolor{white} \cite{radford2021learning}\cite{SurveyonOccupancy} & \cellcolor{white} \cite{wang2024empowering}\cite{SurveyonOccupancy} \\ \hline
Localization & \cellcolor{white} \cite{mohanty2016deepvo}\cite{wang2017deepvo} & \cellcolor{white} \cite{AnImageisWorth}\cite{SurveyonOccupancy} & \cellcolor{white} \cite{SurveyonOccupancy}\cite{erabati2022msf3ddetr} & \cellcolor{white} \cite{SurveyonOccupancy} \\ \hline
Prediction & \cellcolor{white} \cite{shi2024mtr++}\cite{dalcol2024joint}\cite{alahi2016social} & \cellcolor{white} \cite{SurveyonOccupancy}\cite{shi2024mtr++}\cite{dalcol2024joint} & \cellcolor{white} \cite{SurveyonOccupancy}\cite{shi2024mtr++} \cite{alahi2016social} & \cellcolor{white} \cite{SurveyonOccupancy}\cite{dalcol2024joint}\cite{gajewski2024solving} \\ \hline
Planning & \cellcolor{white} \cite{bojarski2016end}\cite{codevilla2018end} & \cellcolor{white} \cite{SurveyonOccupancy}\cite{erabati2022msf3ddetr} & \cellcolor{white} \cite{SurveyonOccupancy}\cite{erabati2022msf3ddetr}\cite{li2022bevformer} & \cellcolor{white} \cite{SurveyonOccupancy}\cite{dalcol2024joint} \\ \hline
Control & \cellcolor{white} \cite{bojarski2016end}\cite{Hwang2024} & \cellcolor{white} \cite{SurveyonOccupancy} & \cellcolor{white} \cite{SurveyonOccupancy}\cite{erabati2022msf3ddetr} & \cellcolor{white} \cite{SurveyonOccupancy}\cite{gajewski2024solving} \\ \hline
\end{tabular}
}
\captionsetup{skip=10pt}
\caption{AI Model Integration Across ADS Layers}
\label{table:ads_layers_comparison}
\end{table}

Table \ref{table:ads_layers_comparison} highlights that legacy AI models remain foundational for real-time tasks on perception, control while FMs drive innovation in planning and perception. VLM's bridge vision to language gaps but are not yet used stand alone in layers such as perception, planning and control. LLMs are well positioned to become the "brain" of autonomous systems, enabling human-like decision making and adaptability in complex scenarios.

\noindent \textbf{Traditional Safety Standards and AI Extensions.} IEC 61508~\cite{IEC61508} and ISO 26262~\cite{ISO26262} offer foundational safety frameworks but lack AI-specific guidance. ISO 21448 (SOTIF)~\cite{ISO21448} extends these for autonomous systems by addressing performance limitations and unexpected scenarios, emphasizing data-related safety measures such as PFMEA for offline training and interpretability analysis to assess ML decisions, but provides limited guidance on architectural safety analyses. AI-focused standards like ISO/PAS 8800~\cite{ISO8800} (automotive) and ISO/IEC TR 5469~\cite{ISO5469} (cross-industry) introduce tailored safety lifecycles, emphasizing data quality, uncertainty quantification, and AI safety properties (robustness, explainability). ISO/PAS 8800~\cite{ISO8800} mandates safety analysis across all development phases and provides informative guidance about existing safety analysis techniques in AI-based systems. The guidance in ISO/TS 5083:2025~\cite{ISO5083} about AI safety largely aligns with ISO/PAS 8800~\cite{ISO8800}. ISO/IEC TS 22440-1~\cite{ISO22440-1} (under development) emphasizes that AI technology selection must be based on safety impact analysis, considering model size, real-time capability, uncertainty estimation, and explainability. ISO/IEC AWI TS 25223~\cite{ISO25223} (under development) focuses on uncertainty quantification methods in AI systems. These standards provide foundational frameworks but lack specific provisions for foundation model characteristics.

\noindent \textbf{Existing AI Safety Analysis Approaches.} FMEA-AI \cite{FMEA-AI} adapts FMEA for AI fairness analysis, applying it to cases like pedestrian and person detection, but lacks architectural analysis. \cite{FTA-FMEAMartinez} reviews FMEA/FTA applications in AI, finding most focus on domain-specific faults rather than analysis of AI-intrinsic failures or AI-specific assets or risks, without conducting new analyses. \cite{FailureModesAerospace} identifies ML-specific errors and conducts a qualitative hazard analysis and FMEA for a visual navigation system, recommending high-level safety requirements without detailed component analysis. \cite{mylius2025systematichazardanalysisfrontier} focuses on system-level hazards in socio-technical systems, including frontier AI, but is less suited for identifying failure modes intrinsic to complex, opaque foundation models.

\noindent \textbf{Fundamental Limitations of Traditional Safety Analysis for AI Systems.} Traditional safety analysis methods (FMEA and FTA) were developed for systems with clear component boundaries and well-understood failure mechanisms. From first principles, these methods are fundamentally limited when applied to AI/FM-based systems in three critical aspects: (1) Abstraction Incompatibility: FMEA and FTA methods operate at component or functional levels with discrete failure states, whereas AI and FM-based systems work with distributed representations in continuous latent spaces where a "failure" is manifested as statistical deviation rather than binary state transitions. (2) Causal Opacity: FMEA and FTA assume traceable cause-effect relationships between events which do not exist in AI/DNN/FM where millions of parameters interact and there are no clear causal chains. (3) Temporal Dynamism: Traditional analyses assume stable system behaviors over time, while AI/DNN/FM-based systems exhibit context-dependent behaviors that change based on input sequences and environmental/ODD factors. These fundamental limitations necessitate new approaches that can address the unique characteristics of foundation model-based systems while maintaining compatibility with established safety engineering practices.

\section{Augmenting Safety Analysis with HySafe-AI}

HySAFE-AI directly addresses the identified limitations by establishing architectural transparency as a fundamental precondition and applying two methodological additions:

\emph{Precondition - Architectural Transparency:} Our safety analysis approach requires sufficient visibility into FM system architecture across different layers and representational dimensions. While FMs remain operationally "closed-box," systematic safety analysis necessitates access to architectural entities that enable identification of failure propagation paths.

\emph{Multi-level Abstraction:} We systematically analyze FM-based systems across architectural entities (from raw inputs to latent spaces), enabling safety analysis to trace failure propagation across system boundaries.

\emph{AI-Specific Failure Taxonomy:} We establish a systematic mapping between standard FMEA guidewords (e.g., “incorrect value,” “missing value”) \cite{SWFMEATech} \cite{SWFMEAfor26262} and generalized \underline{AI Failure Modes} (e.g., hallucination, distributional shift, quantization effects). These AI Failures Modes are broadly applicable across domains, while \underline{Domain-Specific Manifestation} captures how they concretely appear in a given context i.e. in this case autonomous driving.  For example, the general AI failure mode "Hallucination" manifests specifically as "Non-existent objects or agents (e.g., obstacles/pedestrians) in predicted images" within autonomous driving systems. Similarly, "Temporal Reasoning Failure" manifests as "Mispredicted motion of other road users" in the automotive domain. Similarly, quantization effects — a general AI failure mode resulting from the reduction of model precision for hardware efficiency — can be mapped to "missing value" FMEA guideword which in the AD domain manifests as "road boundaries in unusual lighting conditions might not be properly detected".

Table \ref{FMEA_Example} in column "AI Failure Mode (FMEA Guideword)" systematically maps these and other AI-specific failure modes to their corresponding traditional FMEA guidewords. This mapping ensures that established safety frameworks remain applicable even if the underlying technology evolves. By systematically extending the FMEA framework to include these AI-specific failure modes and associating them with established guidewords, HySAFE-AI enables a structured and interpretable risk assessment that bridges AI model behavior with traditional safety engineering practices.

In traditional Fault Tree Analysis (FTA), an E2E AI component would typically be represented as a single node in the fault tree diagram, reflecting its "closed-box" nature. HySAFE-AI extends the traditional FTA by incorporating failure paths that represent latent space errors, temporal mispredictions, etc, enabling the identification failure scenarios that arise from the different blocks of the E2E stack and the interactions within the E2E stack. 

Through this augmented FMEA/FTA methodology, HySAFE-AI not only preserves the rigor of classical safety analysis but also addresses the scale, complexity, and opacity of AI systems, and avoids redundant safety measures ensuring that FM-based autonomous stacks remain robust, dependable, and aligned with evolving safety standards.

\begin{table*}[t]
\centering
{\scriptsize 
\begin{tabular}{|p{0.10\textwidth}|p{0.12\textwidth}|p{0.12\textwidth}|p{0.12\textwidth}|p{0.12\textwidth}|p{0.01\textwidth}|p{0.01\textwidth}|p{0.01\textwidth}|p{0.02\textwidth}|}

\hline
\textbf{System Element} & \textbf{AI Failure Mode (FMEA Guideword)} & \textbf{Domain-Specific Manifestation} & \textbf{Effect} & \textbf{Caused By} & \textbf{S} & \textbf{O} & \textbf{D} & \textbf{RPN} \\
\hline

Latent Denoiser - Quantized Activations & Quantization-Induced Hallucination\break (Incorrect Value) & Imaginary obstacles on driving path, phantom pedestrians crossing highway & False object/obstacle detection & Limited latent feature representation range, noise prediction inaccuracy for fine details & 9 & 7 & 4 & 252 \\ \hline 

Causal Temporal Attention & Temporal Reasoning Failure\break(Incorrect Value/Timing) & Mispredicted motion of other road users in adverse weather conditions & Incorrect velocity/trajectory estimation of vehicles, pedestrians & Traffic scenarios different from training data (heavy snowstorms, dust storms affecting visibility) & 9 & 7 & 4 & 252 \\ \hline

Training Dataset & Dataset Staleness\break(Missing Value) & Missing new road participants (e-scooters, new bike variants) in model awareness & Model fails to adapt to evolving road environment & Outdated training dataset lacking recent traffic participants and road regulations & 9 & 6 & 4 & 216 \\ \hline

Latent Denoiser & Hallucination\break(Incorrect Value) & Non-existent objects or agents in predicted images - imaginary obstacles, phantom pedestrians & False object and obstacle detection & Limited training data, OOD input mapped to nearest known distribution & 9 & 6 & 3 & 162 \\ \hline

UNet - Quantized Weights & Quantization-Induced Feature Extraction Degradation\break(Incorrect or Missing Value) & Road boundaries undetected in unusual lighting conditions & Critical features missed or misclassified & Reduced representational capacity across network layers, diminished generalization & 10 & 5 & 3 & 150 \\ \hline

Trajectory Planner & Constraint Adherence Failure\break(Value too high/low) & Physically impossible maneuvers (90-degree turn at highway speeds, emergency stops) & Vehicle instability, loss of control & No physical constraints embedded in model architecture & 10 & 5 & 3 & 150 \\ \hline

Text Conditioning & Semantic Misinterpretation\break(Incorrect Value) & Wrong action execution in left-hand driving countries, ambiguous intersection commands & Wrong action execution & OOD scenarios, training data bias, ambiguous command interpretation & 10 & 5 & 2 & 100 \\ \hline

Dataset & Data Corruption\break (Incorrect Value) & Systematic prediction biases across driving scenarios & Systematic prediction biases & Label errors, wrong calibration data, sensor miscalibration & 10 & 4 & 2 & 80 \\ \hline
\end{tabular}
}
\captionsetup{skip=10pt}
\caption{HySAFE-AI FMEA with AI-Specific Failure Modes and FMEA Guidewords (S-Severity, O-Occurrence, D-Detection; Values 1–10; RPN = S×O×D)}
\label{FMEA_Example}
\end{table*}

\noindent \textbf{Reference Architecture Use Case: HySafe-AI Walkthrough.}
The following FMEA analysis identifies critical AI-specific failure modes including hallucinations and temporal mispredictions. While not exhaustive, these examples (RPN 100–252) highlight key vulnerabilities in foundation-model-based autonomy without dedicated safety measures, incorporating failure modes from performance optimizations like quantization. The RPN values are based on expert judgment and and the reasoning for the top 3 RPN values is as follows: The \textit{Latent Denoiser - Quantized Activations} component has the highest RPN (252) due to high severity (S=9), as false object detection may trigger emergency maneuvers, risking severe accidents. Its high occurrence (O=7) reflects frequent issues with out-of-distribution inputs. Detection difficulty (D=4) is notable, as quantization artifacts can produce plausible but non-existent objects, making detection challenging. While the \textit{Latent Denoiser} (D=3) is slightly easier to detect, it poses risks before influencing decisions. The \textit{Causal Temporal Attention}, fails when the system mispredicts road user behavior. Severity (S=9) is high since incorrect trajectory estimation can cause collisions. Occurrence (O=7) is due to frequent challenges in unusual conditions. Detection (D=4) is difficult, as distinguishing between reasonable and faulty motion predictions requires complex validation. The \textit{Training Dataset - Staleness} failure (RPN 216) has severity 9, as missing new road users (e.g., e-scooters, new types/shapes of bikes) can lead to severe incidents. Occurrence (O=6) is moderate, since adapting to evolving road environments takes time. Detection (D=4) is hard because it’s difficult to isolate performance degradation caused by dataset staleness versus adversarial inputs.

\begin{figure*}
\centering
\includegraphics[width=6.0in]{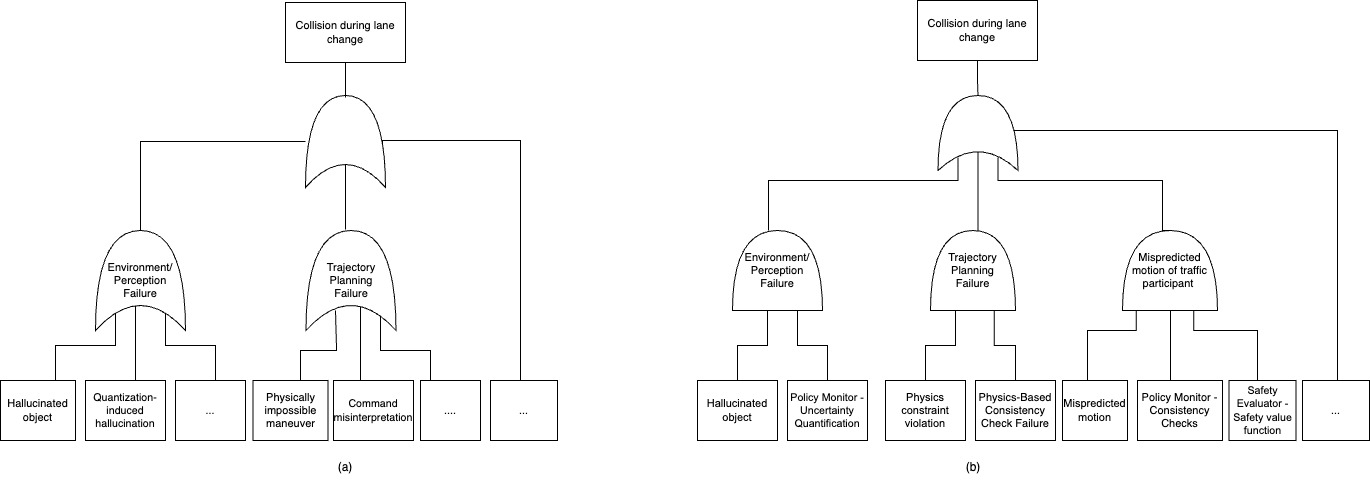}
\caption{(a) HySafe-AI FTA, (b) HySafe-AI FTA with mitigations} \label{combined}
\end{figure*}

The FTA diagram in Figure \ref{combined}(a) provides a partial illustration of how perception and motion planning failures combine to cause unsafe outcomes, highlighting logical dependencies and single points of failure in the E2E reference architecture.

\noindent \textbf{HySafe-AI Outcome: Fused Stack Model.}
The identified high RPN values in the reference architecture FMEA and the identified FTA failures, require architectural safety mitigation which are shown in Table \ref{tab:findings_mitigations}. Moreover, The FMEA in Table \ref{FMEA_with_Mitigations_Example} illustrates how the AI/ML-specific failure modes of the reference architecture can be systematically mitigated through the architectural measures, i.e. safety-aware components like the Policy Monitor and Safety Evaluator, resulting in significantly reduced RPN values. This partial analysis demonstrates, through selected examples, how traditional safety mechanisms (e.g., redundancy, plausibility checks) can be adapted to address AI/ML failures while preserving the predictive capabilities of FMs. The FTA diagram in Figure \ref{combined} (b) illustrates how the AI/ML-specific failure modes of the reference architecture can be systematically mitigated through the architectural measures i.e. safety-aware components like the Policy Monitor and Safety Evaluator, resulting in single point failure becoming dual or multi-point failure. 

\begin{figure}
\centering
\includegraphics[width=3.5in]{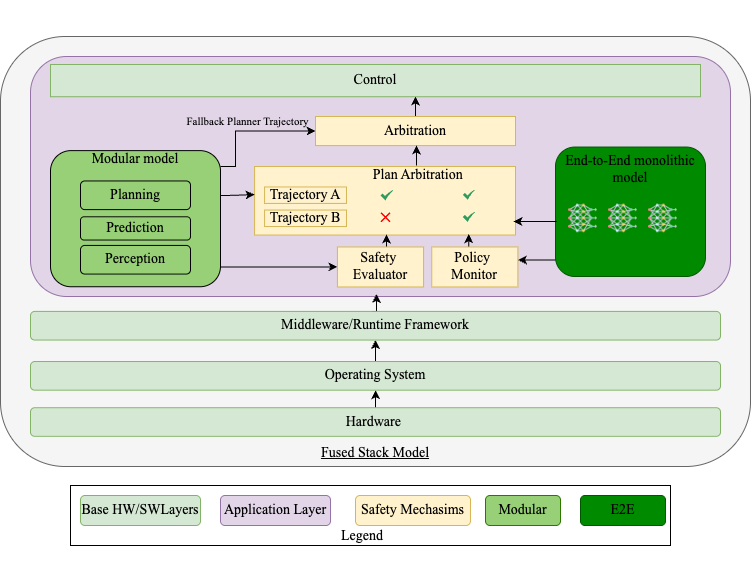}
\caption{Fused Architecture} \label{fig5}
\end{figure}

After the mitigations in Table \ref{tab:findings_mitigations} are applied the resultant architecture is shown in Figure \ref{fig5}. The Policy Monitor evaluates E2E planner reliability via OOD detection and uncertainty quantification, based on level of confidence. The Safety Evaluator applies rule-based and physics-derived checks to reject unsafe trajectories. Plan Arbitration selects the highest-confidence trajectory passing checks both from policy monitor and safety evaluator. Unlike reactive path-tracking systems where there can be comparison of trajectories, this approach pre-validates plans using combined AI/physics safeguards.

\begin{table}
    \centering
    {\scriptsize 
    \begin{tabular}
     {|>{\raggedright\arraybackslash}p{1.8cm}|>{\raggedright\arraybackslash}p{2.4cm}|>{\raggedright\arraybackslash}p{3.4cm}|}
        \hline
        \textbf{FMEA/FTA Finding} & \textbf{Architectural Mitigation} & \textbf{Comment} \\ \hline
        Hallucinated objects & Policy Monitor - Neural Uncertainty Quantification & Detects out-of-distribution and uncertain predictions \\ \hline
        Quantization-Induced Hallucinations & Policy Monitor - Quantization-Calibrated Uncertainty & Uncertainty estimation specifically calibrated for quantized model behavior \\ \hline
        Quantization-Induced Feature Extraction Degradation & Mixed-Precision Architecture & Critical components at INT8 while others at INT4/FP4 \\ \hline
        Quantization-Induced Feature Extraction Degradation & Quantization-Aware Training (QAT) & Simulates quantization effects during training. Reduces accuracy degradation compared to post-training quantization methods. \\ \hline
        Temporal Reasoning Failure & Policy Monitor - Consistency Checks & Validates temporal coherence of predictions \\ \hline
        Constraint Adherence Failure & Safety Evaluator - Physics based consistency checks & Applies vehicle dynamics constraints to trajectories \\ \hline
        Semantic Misinterpretation & Policy Monitor - Learned E2E Verifier & Validates trajectory alignment with commands \\ \hline
        No mechanism to select safe path & Arbitrator - Plan Selection Logic & Selects trajectories with positive verification from multiple monitors \\ \hline
        Dataset Staleness & Active Learning Pipeline \& Over the air update of the model & Continuously updates training data with edge-case models from inference, ensuring the updated model does not perform worse by the re-update of the model \\ \hline
        Corrupted training data & Data Sanitization & Various measures e.g., label consistency validation, sensor calibration validation \\ \hline
    \end{tabular}
    }
    \captionsetup{skip=10pt}
    \caption{Summary of FMEA/FTA Findings and Mitigations}
    \label{tab:findings_mitigations}
\end{table}

\begin{table}
\centering
{\scriptsize 
\begin{tabular}
{|>{\raggedright\arraybackslash}p{1.2cm}|>{\raggedright\arraybackslash}p{2.5cm}|>{\raggedright\arraybackslash}p{1.5cm}|>{\raggedright\arraybackslash}p{0.5cm}|>{\raggedright\arraybackslash}p{0.8cm}|}
\hline
\textbf{System Element} & \textbf{AI Failure Mode\break (FMEA Guideword)} & \textbf{Mitigation Strategy} & \textbf{D Delta} & \textbf{RPN Delta} \\
\hline

Latent Denoiser - Quantized Activations & Quantization induced hallucinations\break (Incorrect Value) & Quantization-aware Policy Monitor & 4→1\break (-3) & 252→63\break (-189) \\ \hline 

Causal Temporal Attention & Temporal Reasoning Failure \break (Incorrect Value/Timing) & Trajectory Consistency Checks & 4→1\break (-3) & 252→63\break (-189) \\ \hline

Training Dataset & Dataset Staleness\break (Missing Value) & Active Learning Pipeline & 4→1\break (-3) & 216→54\break (-162) \\ \hline

Latent Denoiser & Hallucinated objects\break (Incorrect Value) & Neural Uncertainty Quantification & 3→1\break (-2) & 162→54\break (-108) \\ \hline

UNet - Quantized Weights & Quantization-Induced Feature Extraction Degradation \break (Missing Value) & Quantization-Aware Training & 3→1\break (-2) & 150→50\break (-100) \\ \hline

Trajectory Planner & Constraint Adherence Failure\break (Value too high/low) & Safety Value Function & 3→1\break (-2) & 150→50\break (-100) \\ \hline

Text Conditioning & Semantic Misinterpretation\break (Incorrect Value) & E2E Verifier & 2→1\break (-1) & 100→50\break (-50) \\ \hline

Dataset & Corrupted training dataset\break (Incorrect Value) & Data Sanitization & 2→1\break (-1) & 80→40\break (-40) \\ \hline

\end{tabular}
}
\caption{HySafe-AI FMEA with Mitigations (D, RPN: Negative deltas show improvement)}
\label{FMEA_with_Mitigations_Example} 
\end{table}

\section{Conclusion and Future Work}

E2E AI models in ADS face key challenges, including interpretability issues, validation complexity, causal confusion, and latent errors. To address these, the paper proposed HySAFE-AI, a hybrid safety analysis method that adapts traditional FMEA/FTA methods to AI-specific risks. By integrating uncertainty quantification, quantization-aware training, and real-time policy monitors, the risk of safety critical failure risks is reduced. The fused architecture combines foundational models with safety evaluators, ensuring robustness while aligning with guidance from functional safety standards like ISO/PAS 8800 \cite{ISO8800} and ISO 26262 \cite{ISO26262}.
This fused architecture enhances safety in E2E ADS by integrating run-time safety mechanisms like policy monitors (based on uncertainty quantification), safety evaluator and arbitrator etc. However, it introduces computational overhead, potentially impacting latency, a critical factor in time-sensitive driving scenarios. 

Future work will focus on: a) comprehensive analysis of E2E ADS failure modes, and b) recommending design- or runtime mitigation measures based on hazards \cite{singh2021impact} to enhance FM safety in ADS. Collaboration with standards bodies will be key to developing guidance on hybrid-AI safety architectures as technical reports or specifications.

%
%
%
\bibliographystyle{splncs04}

\bibliography{references} 

\begin{thebibliography}{10}
\providecommand{\url}[1]{\texttt{#1}}
\providecommand{\urlprefix}{URL }
\providecommand{\doi}[1]{https://doi.org/#1}

\bibitem{alahi2016social}
Alahi, A., Goel, K., Ramanathan, V., Robicquet, A., Fei-Fei, L., Savarese, S.: Social lstm: Human trajectory prediction in crowded spaces. In: Proceedings of the IEEE Conference on Computer Vision and Pattern Recognition (CVPR). pp. 961--971 (2016)

\bibitem{VisualBackProp}
Bojarski, M., Choromanska, A., Choromanski, K., Firner, B., Jackel, L., Muller, U., Zieba, K.: Visualbackprop: Visualizing cnns for autonomous driving. In: IEEE Conference on Computer Vision and Pattern Recognition (CVPR) Workshops. pp. 470--477 (2017)

\bibitem{bojarski2016end}
Bojarski, M., Del~Testa, D., Dworakowski, D., Firner, B., Flepp, B., Goyal, P., Jackel, L., Monfort, M., Muller, U., Zhang, J., Zhang, X., Zhao, J., Zieba, K.: End to end learning for self-driving cars (2016), available: \url{https://arxiv.org/abs/1604.07316}

\bibitem{E2EDetTransf}
Carion, N., Massa, F., Synnaeve, G., Usunier, N., Kirillov, A., Zagoruyko, S.: End-to-end object detection with transformers. In: Vedaldi, A., Bischof, H., Brox, T., Frahm, J. (eds.) ECCV 2020. LNCS, vol. 12346. pp. 213--229. Springer, Cham (2020)

\bibitem{codevilla2018end}
Codevilla, F., Müller, M., López, A., Koltun, V., Dosovitskiy, A.: End-to-end driving via conditional imitation learning. In: Proceedings of the IEEE International Conference on Robotics and Automation (ICRA). pp. 4693--4700 (2018). \doi{10.1109/ICRA.2018.8460487}

\bibitem{dalcol2024joint}
Dal’Col, L., Oliveira, M., Santos, V.: Joint perception and prediction for autonomous driving: A survey (2024), available: \url{https://arxiv.org/abs/2412.14088}

\bibitem{AnImageisWorth}
Dosovitskiy, A., Beyer, L., Kolesnikov, A., Weissenborn, D., Zhai, X., Unterthiner, T., Dehghani, M., Minderer, M., Heigold, G., Gelly, S., Uszkoreit, J., Houlsby, N.: An image is worth 16x16 words: Transformers for image recognition at scale. In: International Conference on Learning Representations (ICLR) (2021)

\bibitem{erabati2022msf3ddetr}
Erabati, G., Araujo, H.: Msf3ddetr: Multi-sensor fusion 3d detection transformer for autonomous driving (2022), available: \url{https://arxiv.org/abs/2210.15316}

\bibitem{gajewski2024solving}
Gajewski, P., Żurek, D., Pietroń, M., Faber, K.: Solving multi-goal robotic tasks with decision transformer (2024), available: \url{https://arxiv.org/abs/2410.06347}

\bibitem{SWFMEATech}
Goddard, P.: Software fmea techniques. In: Annual Reliability and Maintainability Symposium. 2000 Proceedings. International Symposium on Product Quality and Integrity (Cat. No.00CH37055). pp. 118--123 (2000). \doi{10.1109/RAMS.2000.816294}

\bibitem{FailureModesAerospace}
Hogen, A., Annighoefer, B., Daw, Z.: Failure modes or not failure modes? integrating machine learning in aerospace safety assessment processes (2024). \doi{10.1109/DASC62030.2024.10749313}

\bibitem{Wayve2024}
Hu, A., Russell, L., Yeo, H., Murez, Z., Fedoseev, G., Kendall, A., Shotton, J., Corrado, G.: Gaia-1: A generative world model for autonomous driving (2023), available: \url{https://wayve.ai/research/GAIA-1/}

\bibitem{Hwang2024}
Hwang, J., Xu, R., Lin, H., Hung, W., Ji, J., Choi, K., Huang, D., He, T., Covington, P., Sapp, B., Zhou, Y., Guo, J., Anguelov, D., Tan, M.: Emma: End-to-end multimodal model for autonomous driving. arXiv.org  (October 30 2024), available: \url{https://arxiv.org/abs/2410.23262v1}

\bibitem{IEC61508}
{International Electrotechnical Commission (IEC)}: IEC 61508:2010 -- Functional Safety of Electrical/Electronic/Programmable Electronic Safety-related Systems (2010)

\bibitem{ISO26262}
{International Organization for Standardization}: ISO 26262:2018 -- Road Vehicles – Functional Safety (2018), available: \url{https://www.iso.org/standard/68383.html}

\bibitem{ISO21448}
{International Organization for Standardization}: ISO 21448:2022 -- Road Vehicles — Safety of the Intended Functionality (SOTIF) (2022), available: \url{https://www.iso.org/standard/77490.html}

\bibitem{ISO5469}
{International Organization for Standardization}: ISO/IEC TR 5469:2024 -- Artificial intelligence — Functional Safety and AI systems (2024), available: \url{https://www.iso.org/standard/83303.html}

\bibitem{ISO8800}
{International Organization for Standardization}: ISO/PAS 8800:2024 -- Road vehicles — Safety and Artificial Intelligence (2024), available: \url{https://www.iso.org/standard/16012.html}

\bibitem{ISO5083}
{International Organization for Standardization}: ISO/TS 5083:2025 -- Road vehicles — Safety for automated driving systems — Design, verification and validation (2025), available: \url{https://www.iso.org/standard/81920.html}

\bibitem{ISO22440-1}
{International Organization for Standardization}: ISO/IEC AWI TS 22440-1 -- Artificial intelligence — Functional safety and AI systems, Part 1: Requirements (Under Development), available: \url{https://www.iso.org/standard/89535.html}

\bibitem{ISO25223}
{International Organization for Standardization}: ISO/IEC AWI TS 25223 -- Artificial intelligence — Guidance and requirements for uncertainty quantification in AI systems (Under Development), available: \url{https://www.iso.org/standard/89475.html}

\bibitem{SWFMEAfor26262}
Kim, H.H.: Sw fmea for iso-26262 software development. In: 2014 21st Asia-Pacific Software Engineering Conference. vol.~2 (2014). \doi{10.1109/APSEC.2014.85}

\bibitem{FMEA-AI}
Li, J., C.M.: Fmea-ai: Ai fairness impact assessment using failure mode and effects analysis. In: AI Ethics, Volume 2, 2022. Springer (2022)

\bibitem{li2022bevformer}
Li, Z., Wang, W., Li, H., Xie, E., Sima, C., Lu, T., Qiao, Y., Dai, J.: Bevformer: Learning bird's-eye-view representation from multi-camera images via spatiotemporal transformers. In: Computer Vision – ECCV 2022: 17th European Conference on Computer Vision, Tel Aviv, Israel, October 23–27, 2022, Proceedings, Part IX. pp. 1--17. Springer (2022)

\bibitem{FTA-FMEAMartinez}
Martinez, J., Eguia, A., Urretavizcaya, I., Amparan, E., Negro, P.L.: Fault tree analysis and failure modes and effects analysis for systems with artificial intelligence: A mapping study. In: 2023 7th International Conference on System Reliability and Safety (ICSRS) (2023). \doi{10.1109/ICSRS59833.2023.10381456}

\bibitem{mohanty2016deepvo}
Mohanty, V., Agrawal, S., Datta, S., Ghosh, A., Sharma, V., Chakravarty, D.: Deepvo: A deep learning approach for monocular visual odometry (2016), available: \url{https://arxiv.org/abs/1611.06069}

\bibitem{mylius2025systematichazardanalysisfrontier}
Mylius, S.: Systematic hazard analysis for frontier ai using stpa (2025), \url{https://arxiv.org/abs/2506.01782}

\bibitem{pitale2024}
Pitale, M., Abbaspour, A., Upadhyay, D.: Inherent diverse redundant safety mechanisms for ai-based software elements in automotive applications. arXiv preprint  (2024)

\bibitem{radford2021learning}
Radford, A., Kim, J., Hallacy, C., Ramesh, A., Goh, G., Agarwal, S., Sastry, G., Askell, A., Mishkin, P., Clark, J., Krueger, G., Sutskever, I.: Learning transferable visual models from natural language supervision (2021), available: \url{https://arxiv.org/abs/2103.00020}

\bibitem{rosero2024hybrid}
Rosero, L., Gomes, I., da~Silva, J., Przewodowski, C., Wolf, D., Osório, F.: Integrating modular pipelines with end-to-end learning: A hybrid approach for robust and reliable autonomous driving systems. Sensors (Basel)  (2024). \doi{10.3390/s24072097}

\bibitem{Wayve2025}
Russell, L., Hu, A., Bertoni, L., Fedoseev, G., Shotton, J., Arani, E., Corrado, G.: Gaia-2: A controllable multi-view generative world model for autonomous driving (2025), available: \url{https://wayve.ai/research/GAIA-2/}

\bibitem{shah2022lmnav}
Shah, D., Osinski, B., Krueger, Ichter, B., Levine, S.: Lm-nav: Robotic navigation with large pre-trained models of language, vision, and action (2022), available: \url{https://arxiv.org/abs/2207.04429}

\bibitem{shi2024mtr++}
Shi, S., Jiang, L., Dai, D., Schiele, B.: Mtr++: Multi-agent motion prediction with symmetric scene modeling and guided intention querying. IEEE Transactions on Pattern Analysis and Machine Intelligence  \textbf{46}(5),  3955--3971 (2024). \doi{10.1109/TPAMI.2024.3352811}

\bibitem{singh2021impact}
Singh, V., Pitale, M.: Impact of automotive system safety design on machine learning based perception systems. In: 4th IEEE International Conference on Industrial Cyber-Physical Systems (2021)

\bibitem{wang2017deepvo}
Wang, S., Clark, R., Wen, H., Trigoni, N.: Deepvo: Towards end-to-end visual odometry with deep recurrent convolutional neural networks. In: IEEE International Conference on Robotics and Automation (2017)

\bibitem{wang2024empowering}
Wang, Y., Jiao, R., Zhan, S., Lang, C., Huang, C., Wang, Z., Yang, Z., Zhu, Q.: Empowering autonomous driving with large language models: A safety perspective (2024), available: \url{https://arxiv.org/abs/2403.12345}

\bibitem{E2ELearningApproach}
Wang, Y., Liu, D., Jeon, H., Chu, Z., Matson, E.: End-to-end learning approach for autonomous driving: A convolutional neural network model. Tech. rep., Department of Computer and Information Technology, Purdue University, U.S.A.

\bibitem{SurveyonOccupancy}
Xu, H., Chen, J., Meng, S., Wang, Y., Chau, L.: A survey on occupancy perception for autonomous driving: The information fusion perspective. Information Fusion  \textbf{114},  102671 (2025). \doi{10.1016/j.inffus.2024.102671}

\bibitem{Yang2023B}
Yang, J., Gao, S., Qiu, Y., Chen, L., Li, T., Dai, B., Chitta, K., Wu, P., Zeng, J., Luo, P., Zhang, J., Geiger, A., Qiao, Y., Li, H.: Genad: Generalized predictive model for autonomous driving (2023), available: \url{https://arxiv.org/abs/2403.09630}

\bibitem{Yang2023A}
Yang, Z., Jia, X., Li, H., Ya, J.: Llm4ad: A survey of large language models for autonomous driving (2023), available: \url{https://arxiv.org/abs/2311.01043v1}

\end{thebibliography}

\end{document}